\documentclass[conference]{IEEEtran}

\ifCLASSINFOpdf

\usepackage{mdwmath}
\usepackage{mdwtab}
\usepackage{graphicx}

\begin{document}
%
\title{Adversarial  Domain Adaptation for Action Recognition Around the Clock }

 \author{ \IEEEauthorblockN{Anwaar Ulhaq}
\IEEEauthorblockA{School of Computing, Mathematics and Engineering,\\ Faculty of Business, Justice and Behavioural Sciences,\\ 
Charles Sturt University, NSW, Australia,\\ 
Email:aulhaq@csu.edu}
}


\maketitle

\begin{abstract}
Due to the numerous potential applications in visual surveillance and nighttime driving, recognizing human action in low-light conditions remains a difficult problem in computer vision. Existing methods separate action recognition and dark enhancement into two distinct steps to accomplish this task. However, isolating the recognition and enhancement impedes end-to-end learning of the space-time representation for video action classification. This paper presents a domain adaptation-based action recognition approach that uses adversarial learning in cross-domain settings to learn cross-domain action recognition. Supervised learning can train it on a large amount of labeled data from the source domain (daytime action sequences). However, it uses deep domain invariant features to perform unsupervised learning on many unlabelled data from the target domain (night-time action sequences). The resulting augmented model, named 3D-DiNet can be trained using standard backpropagation with an additional layer. It achieves SOTA performance on InFAR and XD145 actions datasets.
\end{abstract}

\IEEEpeerreviewmaketitle

\section{Introduction}

Recognizing human actions in diverse environments remains a challenging task for computer vision. During the past decade, a large number of automated action recognition approaches that employ various modalities have been developed\cite{1,survey2022,filter,ulhaq2022vision}. Given the limited performance of traditional features and domain-specific machine learning methods, researchers concentrate on deep learning-based techniques that process data from end to end for feature extraction and classification. CNN-based methods extract features hierarchically, with initial layers extracting local features and final layers extracting global features.

However, the majority of the aforementioned methods only apply to daytime videos or when good lighting conditions are present; they do not account for poor lighting conditions or the recognition of actions at night. The lack of research into action recognition in dark videos can be attributed to the following two factors: (i) Ineffective data enhancement techniques, such as image enhancements that could improve the appearance of dark video frames, may not consistently improve the action recognition accuracy of dark videos, (ii) compared to the availability of large action datasets such as Kinetic 400 \cite{Kinetics400} and Kinetic 600 \cite{Kinetics600} for general action recognition, insufficient datasets are available for such an investigation.

\begin{figure}
\centering
\includegraphics[width=70 mm]{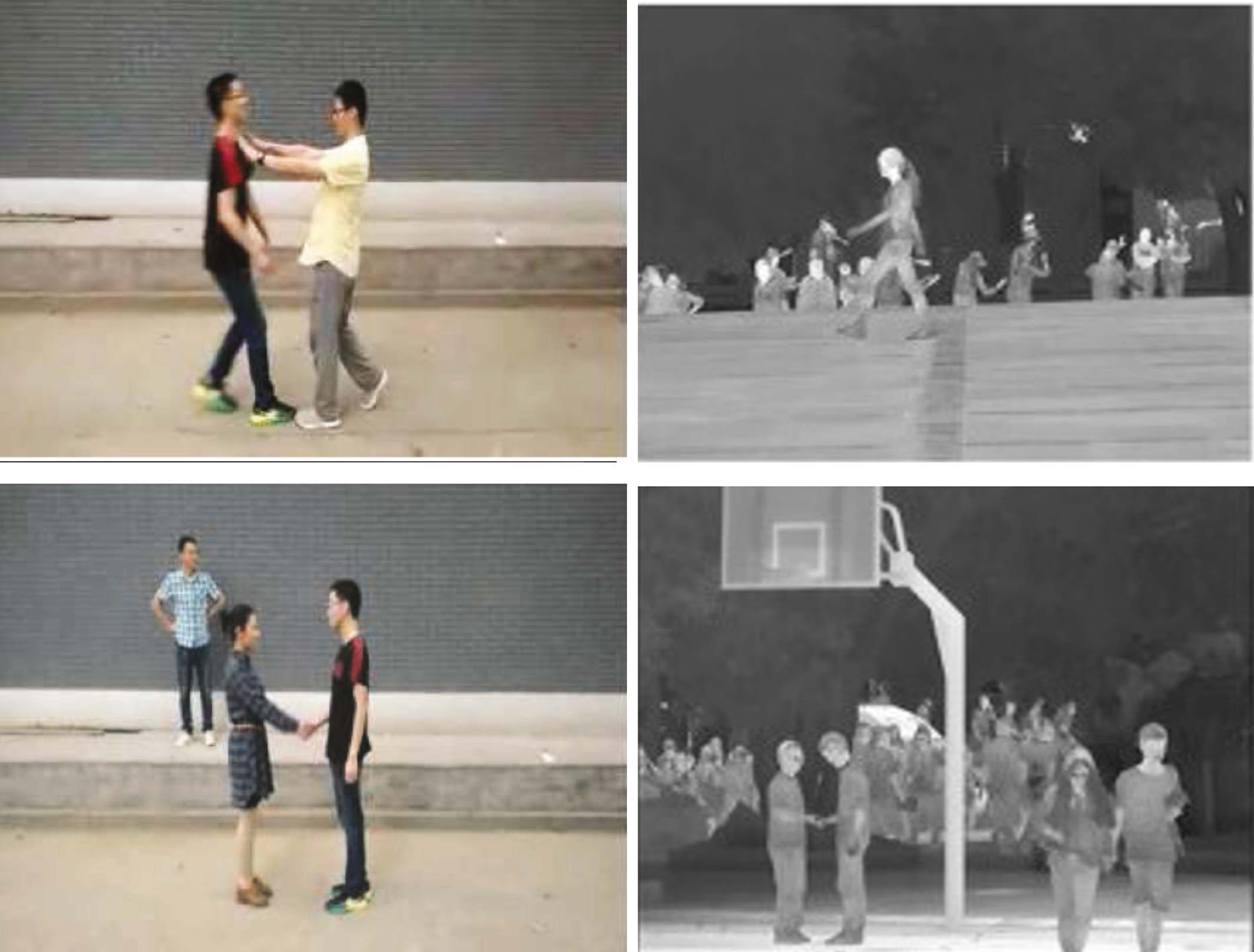} 
\caption{
Different examples of day and night-time actions (Punch, walk, handshaking, hand-shake)-clockwise from two different domains. Actions can easily be captured from the RGB and labeling is easy so it can be used as a source domain while the infrared domain is a good candidate for the target domain. Unsupervised domain adaptation helps to recognize target domain action without any labeling.}
\label{fig:ACCV22}
\end{figure}

Unlike other works that consider night-time action recognition as an image enhancement problem \cite{11,12,ulhaqfilter2,haqvideo}, this paper considers it a domain shift problem. However, despite the deep neural network's extraordinary success in a wide variety of application scenarios, its generalization performance in other new domains remains subpar due to the domain shift problem \cite{Domainshift, DAsurvey}.  Domain adaptation has proved to be a reliable solution to the domain shift problem and various domain adaptation approaches exist in the literature.  A promising direction of research is unsupervised domain adaptation (UDA) \cite{UDAsurvey} which is intended to transfer knowledge from a labeled source domain to an unlabeled target domain.  However, the majority of these approaches work well for images or noisy videos.

This paper extends unsupervised domain adaptation to learn complex action representations across different domains. Thus, we start by learning features that combine (i) feature discrimination and (ii) domain independence. This is accomplished by optimizing the underlying features and two discriminative classifiers operating on these features simultaneously: It can be performed by the action classifier that predicts the action class and is used both during training and at test time and the domain classifier that discriminates between the source and the target domains. It can be done in an adversarial learning network setting to minimize the loss of the action classifier while maximizing the loss of the domain classifier.

The proposed architecture is composed deep feed-forward network that uses standard layers and loss functions of a 3D convolutional neural network,  with a trivial gradient reversal layer that leaves the input unchanged during forward propagation but reverses the gradient by multiplying it by a negative scalar during the backpropagation. This 3D design and gradient reversal layer can also be considered an extension of similar 2D counterparts for image-based adversarial learning for generic tasks \cite{UDAresidual, UDAmain}. However, the proposed model takes care of space-time convolutions and adversarial learning.

Our work is inspired by the success of domain adaptation methodologies to achieve promising results from transfer learning across domains. In summary, our contributions are three-fold:

1-	We propose a 3D convolutional neural network, named 3D-DiNet for day-night action recognition by extending 3D convolutional network architecture by integrating unsupervised domain adaptation to learn domain invariant features. 
2-	To the best of our knowledge, the proposed model is the first of its kind for its domain invariant video-based architecture for action recognition in the dark. 
3-	It achieves considerable performance compared to the state-of-the-art on benchmark datasets.

\subsection{Related Work}

Different challenging scenarios for human action recognition are considered in the past \cite{survey2022} with a large number of potential applications in different areas ( e.g. visual surveillance, video retrieval, sports video analysis, human-computer interfaces, and smart rooms). Different scenarios include but are not limited to action in the wild (YouTube videos and movies), actions in group formations, actions in the crowd, actions across different viewpoints, and in the presence of occlusion \cite{chalenge1,challenge2}.  An interesting challenging scenario is adverse lighting conditions or the recognition of actions at night-time. This work is related to approaches that take action recognition in the dark using different sensor modalities. 

Due to the low quality of dark images, the existing optical flow estimation methods \cite{8} are unable to obtain accurate optical flow for action recognition. Consequently, the two-stream methods \cite{6,7} perform poorly as well. Consequently, image enhancement is frequently regarded as a prerequisite for action recognition tasks. Multi-scale retinex  \cite{9}  provides colour constancy and dynamic range compression by combining multiple SSR outputs into a single output image using a retinex. LIME \cite{LIME} accomplishes the improvement by estimating and refining a low-light image illumination map. However, it is difficult to insolvent an end-to-end network by integrating these approaches. 

Another approach is to fuse multiple spectrums to solve this issue. A series of works have been proposed by \cite{11,12} for simultaneous action recognition from multiple video streams. Recently, Xu et al. \cite{ARID} compiled the first dataset titled ARID that focuses on human actions in dark videos and discovered that current action recognition models and frame enhancement techniques are ineffective for action recognition in the dark. Dark-Light networks \cite{Darklight} utilize both dark videos and their brightened counterparts to form a dual-pathway structure for effective video representation, which greatly improves action recognition performance but is computationally intensive.

In recent years, several domain adaptation techniques have been proposed \cite{DAsurvey}. Domain adaptation is intended to train a neural network on a source dataset and secure a good accuracy on the target dataset which is significantly different from the source dataset. Many methods perform unsupervised domain adaptation by mapping feature distributions in the source and target domains. Chopra et al \cite{chopra} perform the multi-level training of a sequence of deep autoencoders, gradually replacing source domain examples with target domain examples. Yaroslav and Victor \cite{UDAmain} perform feature learning, domain adaptation, and classifier learning jointly, in a unified architecture and using a single learning algorithm (backpropagation). Our proposed approach is closer to their work.

\section{THE PROPOSED METHOD}
We first provide an overview of the architecture of the proposed model in Section A. Then, domain adversarial learning is discussed in the following subsections.

\subsection{The Backbone: 3D convolution-based action recognition: } The straightforward solution to exploit spatio-temporal information is to perform 3D convolution on video which was validated in some pioneering CNN-based action recognition works \cite{CNNreview}. 3D convolution is achieved by convolving a 3D kernel into a video clip. The operation at position (x, y, t) in the jth convolution kernel in the ith layer and the kernel convolutes the mth feature map of the previous layer, then it  is formalized as:
\begin{equation}
    \begin{array}{cc}
  V_{i,j}^{x,y,t} =   & b_{i,j}+\phi(\sum_{m} \\
     & \sum_{h=0}^{Hi-1} \sum_{w=0}^{Wi-1} \sum_{l=0}^{Li-1} \\
     & W_{i,j,m}^{h,w,t}V_{i-1}^{(x+h),(y+w),(t+l)} ),
\end{array}
\end{equation}


Here $\phi$ is a non-linear (e.g., Tanh, Sigmod, or ReLU) activation function, b is the bias, W is the 3D weight matrix, and H, W, and L are the height, width, and temporal length of the kernel, respectively.

\begin{figure*}
\centering
\includegraphics[width=160 mm]{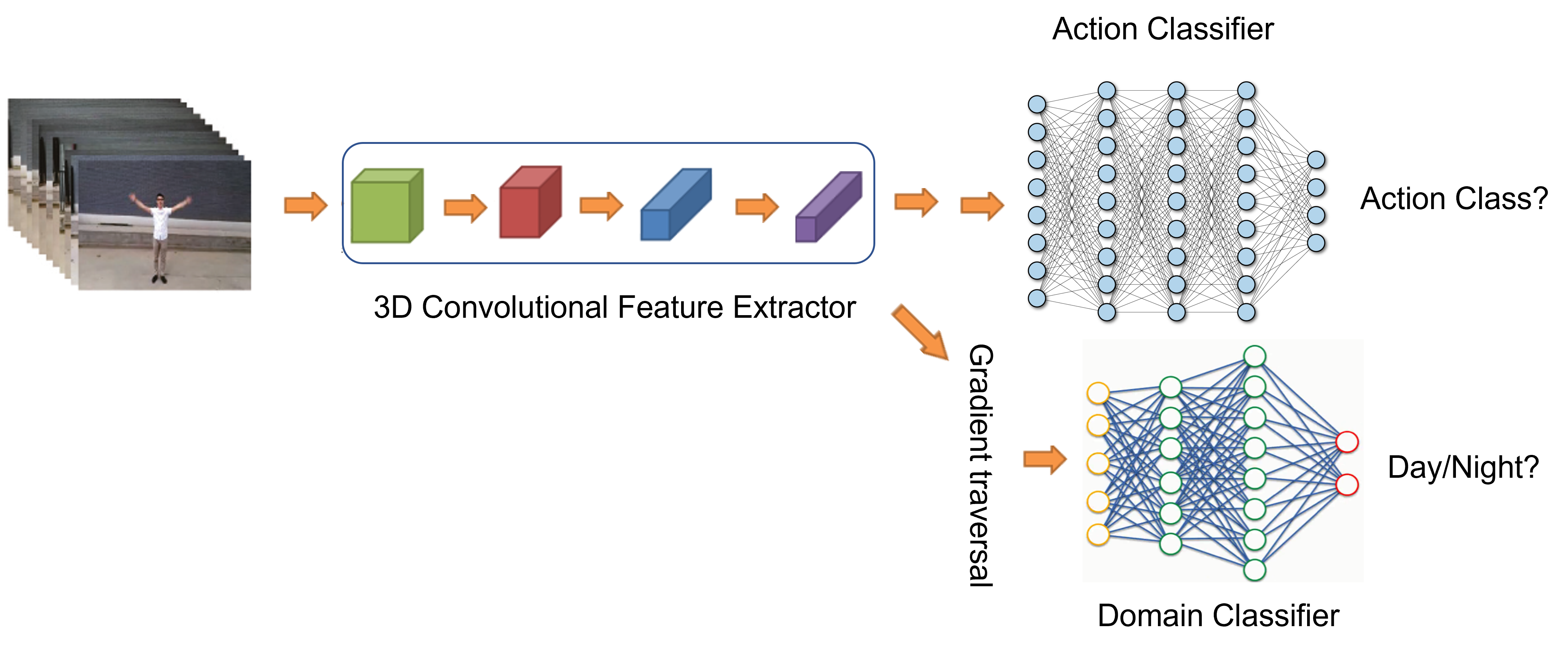} 
\caption{
Architectural diagram of the proposed framework for training purposes. Only the source domain is used for action classifier training. Both source (daytime) and target domain (Night-time) videos are used as input for the domain classifier. At inference time, only one modality (Target domain video) is used.}
\label{fig:DICTA}
\end{figure*}

Various models incorporate 3D convolutions for action recognition. however, we will discuss important variants that can be used as part of our framework. 

\textbf{C3D and Res3D}: Tran et al. \cite{C3D} conducted a systematic study to find the best temporal kernel length for 3D CNN and developed a VGG-style 3D CNN architecture named C3D. The C3D architecture consists of 8 convolutional layers with small $3\times 3\times 3$ convolutional kernels, five pooling layers, and two fully connected layers. The extracted C3D features were demonstrated to be generic, efficient, and compact. Furthermore, Tran et al. \cite{C3Dupdated} conducted a 3D CNN search in a deep residual learning framework and developed a ResNet18-style 3D CNN architecture named Res3D, which outperforms C3D by a good margin in terms of recognition accuracy. In addition, Res3D is 2 times faster in run-time, 2 times smaller in mode size, and more compact than C3D. By pre-training on the largest action recognition benchmark Sports-1 M, the C3D, and Res3D work both provided their pre-trained model, which can be used either as the initialization in transfer learning or as a fixed spatiotemporal feature extractor.

\textbf{ResNeXt-101}: The ResNeXt101 is a model is based on regular ResNet model, substituting 3x3 convolutions. 
Unlike the original bottleneck block, the ResNeXt block introduces group convolutions, which divide the feature maps into small groups. ResNeXt introduces cardinality, which is a different dimension from deeper and wider.  Carnality refers to the number of middle convolutional layer groups in the bottleneck block. We use  ResNeXt-101 using the carnality of 32. In this paper, we fined-tuned ResNeXt-101 \cite{ConNEXT} pre-trained on Kinetic -400 as a backbone for the proposed 3D-DiNet (3D-Domain -invariant Network) architecture. 

\begin{figure*}
\centering
\includegraphics[width=160 mm]{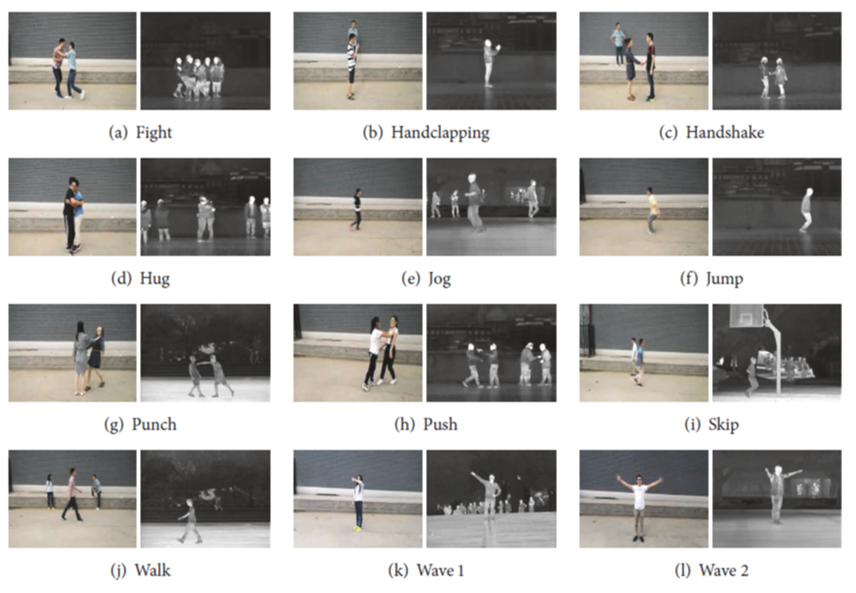} 
\caption{
Visible and infrared actions, respectively. On the left of each subfigure is the visible image from the XD145 dataset's video sequences, while on the right is the infrared image from the InFAR dataset's video sequences.}
\label{fig:ACCV22}
\end{figure*}

\textbf{Learning Domain-invariant Features:} Assume that the model works with input samples $x \in X$, where X is some input space and predicts action labels (output) y from the label space Y. There exist two distributions S(x, a)and T(x, a), the source distribution (in other words, S is“shifted” from T by some domain shift). If we can find a mapping $M_{f}, f=M_{f}(x,\theta_{f})$ where $\theta_{f}$ is the vector of parameters of layers of a feed-forward network  that extracts feature vector f, our goal is to find domain-invariant features such that:

\begin{equation}
    S(f)={M_{f}(x,\theta_{f})|x\sim S(x)}
\end{equation}
and,
\begin{equation}
    T(f)={M_{f}(x,\theta_{f})|x\sim T(x)}
\end{equation}

 become similar. We can divide the network into three parts: a feature extractor with parameters  $\theta_{f}$ that extracts features, a fully connected network with parameters  $\theta_{a}$ that takes f as input and maps it to an action class labels, and a domain classifier with parameters  $\theta_{d}$ that takes f as input and determines domain labels as night or day actions. Therefore, to extract domain invariant features, we need $\theta_{f}$ that jointly minimizes the loss of the action classifier and domain classifier. In other words, we are seeking a functional E such that:
 
\begin{equation}
    \begin{array}{cc}
 F(\theta_{f},\theta_{a},\theta_{d}) =   & \sum_{i=1,...,N,di=0}L_{a}^{i} \\
     & (\theta_{f},\theta_{a})-\lambda \\
     & \sum_{i=1,...,N}L_{d}^{i}(\theta_{f},\theta_{d}),
\end{array}
\end{equation}

At its saddle point, the parameters of the domain classifier will minimize domain shift while minimizing action classification loss. The parameter $\lambda$ controls the trade-off between two objectives. The stochastic updates will be as follow:

\begin{equation}
\theta_{f} \leftarrow \theta_{f} -\alpha (\frac{\partial L_{a}^{i}}{\partial \theta_{f}} - \lambda \frac{\partial L_{d}^{i}}{\partial \theta_{f}})
  \label{equ:dt}
\end{equation}

\begin{equation}
\theta_{a} \leftarrow \theta_{f} -\alpha (\frac{\partial L_{a}^{i}}{\partial \theta_{f}} )
  \label{equ:dt}
\end{equation}

\begin{equation}
\theta_{d} \leftarrow \theta_{f} -\alpha ( \frac{\partial L_{d}^{i}}{\partial \theta_{f}})
  \label{equ:dt}
\end{equation}

A tricky solution to this problem can be implemented during backpropagation by inverting the effect of domain classification loss.  It gradient from the subsequent layers is multiplied with $-\lambda$ and passed to the preceding layers.  Such operation can be implemented in a separate layer between the feature extractor and domain classifier that acts as an identity layer during forward pass while reversing the gradient during back-propagation. It will reduce domain shift by confusing the classifier. Such a layer is successfully implemented by \cite{UDAmain} for domain adaptation in the image domain. 

For the domain adaptor, three fully connected layers were used with binomial cross-entropy loss while the action classifier used logistic regression loss.

\section{Experimental Results} 

\textbf{Datasets:} The InFAR dataset  and the XD145 dataset
(the dataset will be available at: $https://sites.google.com/site/yangliuxdu/)$  are used for the visible-to infrared action recognition task, where the XD145 dataset is used as the source domain and the InFAR dataset is used as the target domain. 

(A) InFAR
The InFAR dataset \cite{InFAR} includes 600 video sequences captured by infrared thermal imaging cameras. As depicted in Figure 3, the dataset consists of the fight, handclapping, handshake, hug, jog, jump, punch, push, skip, walk, wave 1 (one-hand wave), and wave 2 (two-hand wave), where each action class contains 50 video clips with an average duration of 4 seconds.

The frame rate is 25 frames per second, and the resolution is 293 by 256 pixels. Each video depicts single or multiple actions carried out by a single or multiple individuals. Some of them involve interactions between multiple individuals, as configuration details in Figure 3.

\begin{figure*}
\centering
\includegraphics[width=160mm]{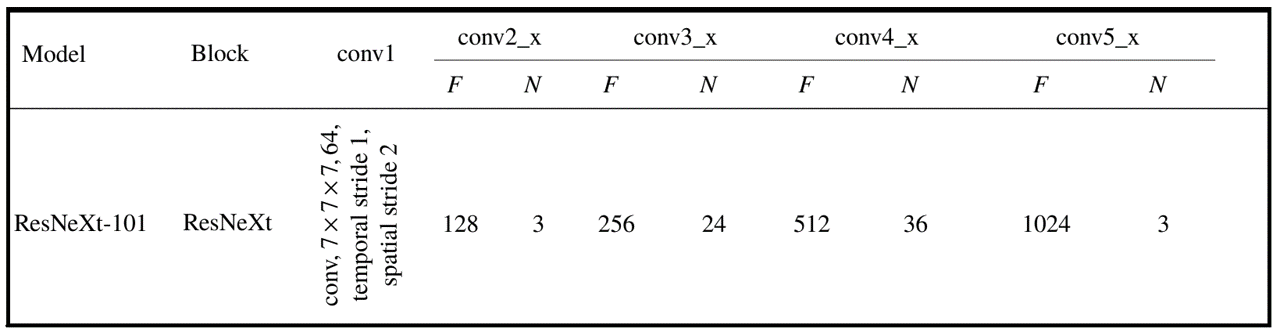} 
\caption{Configuration details for 3D feature extractor: Each convolutional layer is followed by batch normalization and a ReLU. Spatio-temporal down-sampling is performed by conv3-1, conv4-1, and conv5-1 with a stride of 2. F is the number of feature channels and N is the number of blocks in each layer.}
\label{fig:ACCV22}
\end{figure*}

(B) XD145
A visible light action dataset named XD145 is a reflection of the above InFAR dataset. In correspondence with the target domain action categories, both the XD145 and the InFAR datasets have the same action categories, as shown in Figure 3. The XD145 action dataset consists of 600 video sequences captured by visible light cameras, and there are 50 video clips for each action class. All actions were performed by 30 different volunteers. Each clip lasts for 5 s on average. The frame rate is 25 fps, and the resolution is $320 \times 240$. As shown in Figure 4, the background, pose, and viewpoint variations are considered when constructing the dataset to make the dataset more representative of real-world scenarios.

\textbf{The training and Testing Setup}: The framework works in a bipartite manner as follows: First feature extractor and action predictor on source data (daytime action sequences) as labels are available. Second, a domain classifier is trained on both source and target data as action labels are not available. During the learning process and standard backpropagation, the feature extractor will be updated from the loss of both domain classifier and action predictor. As gradients from the domain classifier will be multiplied with a small negative constant, the network will learn domain-invariant features from a domain classifier perspective. Finally, the domain classifier part is removed and the feature extractor and action predictor are used for action predictors for test actions. Since pre-training on large-scale datasets is an effective way to achieve good performance levels on small datasets, we used the deep 3D ResNets pre-trained available at 1https://github.com/kenshohara/3D-ResNets-PyTorch.

\begin{figure}
\centering
\includegraphics[width=70 mm]{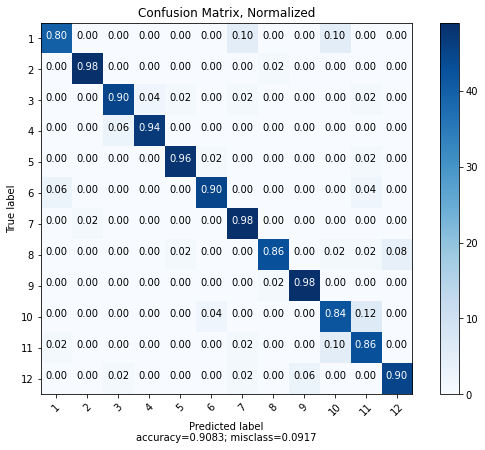} 
\caption{The confusion matrix for 3D-DiNet on InFAR action recognition dataset as test dataset. However, It includes fight, handclap-ping, handshake, hug, jog, jump, punch, push, skip, walk, wave1,  (one-hand wave), and wave 2 (two-hand wave) actions. }
\label{fig:ACCV2}
\end{figure}

The standard configuration of the used feature extractor is shown in Figures 4 and 6. Each convolutional layer is followed by batch normalization and a ReLU. Spatio-temporal down-sampling is performed by conv3-1, conv4-1, and conv5-1 with a stride of 2. F is the number of feature channels and N is the number of blocks in each layer. We represent conv, $x^{3}$, F as the kernel size, and the number of feature maps of the convolutional filter is $x \times x \times x$, group as the number of groups of group convolutions, which divide the feature maps into small groups. BN refers to batch normalization. Shortcut connections of the architectures are shown as summations. For domain classifier training, half of each batch is populated by the samples from the source domain (with known labels), and the rest is comprised of the target domain (with unknown labels). 

\begin{figure}
\centering
\includegraphics[width=70 mm]{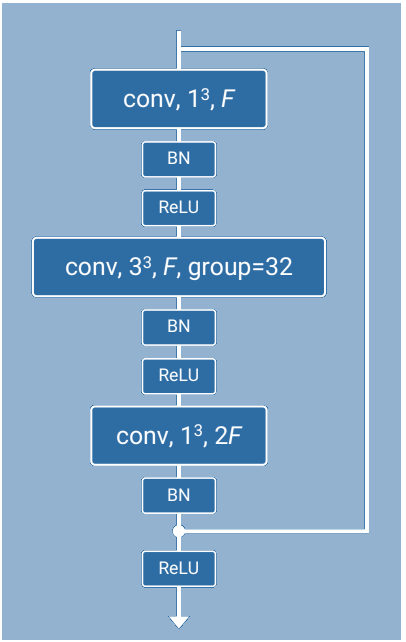} 
\caption{We represent Conv, $x^{3}$, F as the kernel size, and the number of feature maps of the convolutional filter is $x \times x \times x$ and F, respectively, and the group as the number of groups of group convolutions, which divide the feature maps into small groups. BN refers to batch normalization. Shortcut connections of the architectures are shown as summations. }
\label{fig:ACCV225}
\end{figure}

The value of adaptation factor $\lambda$ is controlled gradually than using as a fixed value described by the following formula:

\begin{equation}
\lambda_{p}=\frac{2}{1+exp(-10.p)}-1
  \label{equ:dt}
\end{equation}
where p is the training progress linearly changing from 0 to 1. We use stochastic gradient descent with 0.9 momentum and  the learning rate annealing

\textbf{Comparisons With State-of-The-Art Methods}

We used three backbones for the domain adaptation network and named the resulting networks C3D-DA, ResNet-DA, and ResNeXt -DA. We calculated confusion matrices for proposed variants. Figure 5 displays the confusion matrix for the proposed  3D-DiNet model with an improved average recognition accuracy of 90.83 compared to other networks. 

In Table 1, we have provided the performance of different CNN-based models on the InFAR dataset.  It shows that with the inclusion of the proposed domain adaptation strategy with gradient traversal layer, their performance on the InFAR dataset exceeds by a clear margin. We found that the Top-1 accuracy of our best is increased by a good margin of 6.58 percent, which proves not only the proposed method is powerful, but the optical flow may not be useful for AR in the dark. Adaptation is more successful when the source domain test error is low, while the domain classifier error is high.

\begin{table}[htb]
\caption{Average recognition accuracy results of a few competitive models and ours on InFAR dataset.}
\begin{center}
\begin{tabular}{lll}
\hline
Method $\qquad\qquad$ & Citation & Top-1 Accuracy \\
\hline
iDT & \cite{iDT} & 71.35 \\
2 Stream 2D CNN &  \cite{InFAR}& 76.66 \\
2 Stream 3D CNN &  \cite{C3D}& 77.5 \\
CDFAG  & \cite{InFAR}& 78.55 \\
4-Stream CNN &  \cite{4} & 83.40 \\
3D-ResNext-101 & \cite{3DResnNet101}& 86.36 \\
SCA&  \cite{CSA}& 84.25\\
 3D-DiNet& The proposed & 90.83 \\
 \end{tabular}
 \end{center}
\end{table}

\section{Conclusion}
In this paper, we proposed a domain adaptation-based action recognition model named 3D-DiNet that uses adversarial learning in cross-domain settings to learn cross-domain action recognition. It learns deep domain invariant features to perform unsupervised learning on any unlabelled data from the target domain (night-time action sequences).3D-DiNet is trained on the XD145 actions dataset (daytime actions) and tested on the InFAR action dataset (infra-red actions) and the model achieves SOTA performance on InFAR with a clear margin of 6.58 percent compared to other existing approaches.
In the future, we will extend the concept of domain-variance action recognition to other modalities as well.



\bibliographystyle{IEEEtran}

\end{document}